%% file: colm2024_conference.tex
\documentclass{article} %
\usepackage{colm2024_conference}

\usepackage{booktabs}
\usepackage{graphicx}
\usepackage{enumitem}
\usepackage{wrapfig}
\usepackage{algorithm}
\usepackage{algpseudocode}

\usepackage{microtype}
\usepackage{amsmath}
\usepackage{colortbl}
\usepackage[utf8]{inputenc}
\definecolor{lightgray}{rgb}{0.9,0.9,0.9}
\usepackage{caption}
\usepackage{subcaption}
\usepackage{xcolor}
\usepackage{setspace}
\usepackage{url}
\usepackage{multirow}
\usepackage{colortbl}
\usepackage{tabularx}
\usepackage{blindtext}
\usepackage{pgfplots}
\pgfplotsset{compat=1.18} 
\usepackage{tikz}
\usetikzlibrary{er,positioning,bayesnet}
\usepackage{makecell}
\usepackage{tipa}
\usepackage{siunitx}
\usepackage{nicefrac}
\usepackage{tocloft}
\usepackage{listings}
\usepackage[raster,skins]{tcolorbox} %
\usepackage{xltabular}
\usepackage{adjustbox}
\usepackage{xurl}
\usepackage{cleveref}
\input{math_commands.tex}

\title{TBAC-UniImage: Unified Understanding and Generation by Ladder-Side Diffusion Tuning}

\author{
  \bf Junzhe Xu, Yuyang Yin, Xi Chen\\
  Basic Algorithm Center, PCG, Tencent
}

\begin{document}

\maketitle

\begin{abstract}
This paper introduces TBAC-UniImage, a novel unified model for multimodal understanding and generation. We achieve this by deeply integrating a pre-trained Diffusion Model, acting as a generative ladder, with a Multimodal Large Language Model (MLLM). Previous diffusion-based unified models face two primary limitations. One approach uses only the MLLM's final hidden state as the generative condition. This creates a shallow connection, as the generator is isolated from the rich, hierarchical representations within the MLLM's intermediate layers. The other approach, pretraining a unified generative architecture from scratch, is computationally expensive and prohibitive for many researchers. To overcome these issues, our work explores a new paradigm. Instead of relying on a single output, we use representations from multiple, diverse layers of the MLLM as generative conditions for the diffusion model. This method treats the pre-trained generator as a ladder, receiving guidance from various depths of the MLLM's understanding process. Consequently, TBAC-UniImage achieves a much deeper and more fine-grained unification of understanding and generation. 
% Our model achieves promising results on three standard benchmarks, demonstrating the effectiveness of this deep-integration strategy.
\end{abstract}

\vspace{-15pt}

\begin{figure}[hbp]
    \centering
    \includegraphics[width=0.85\textwidth, height=13.2cm]{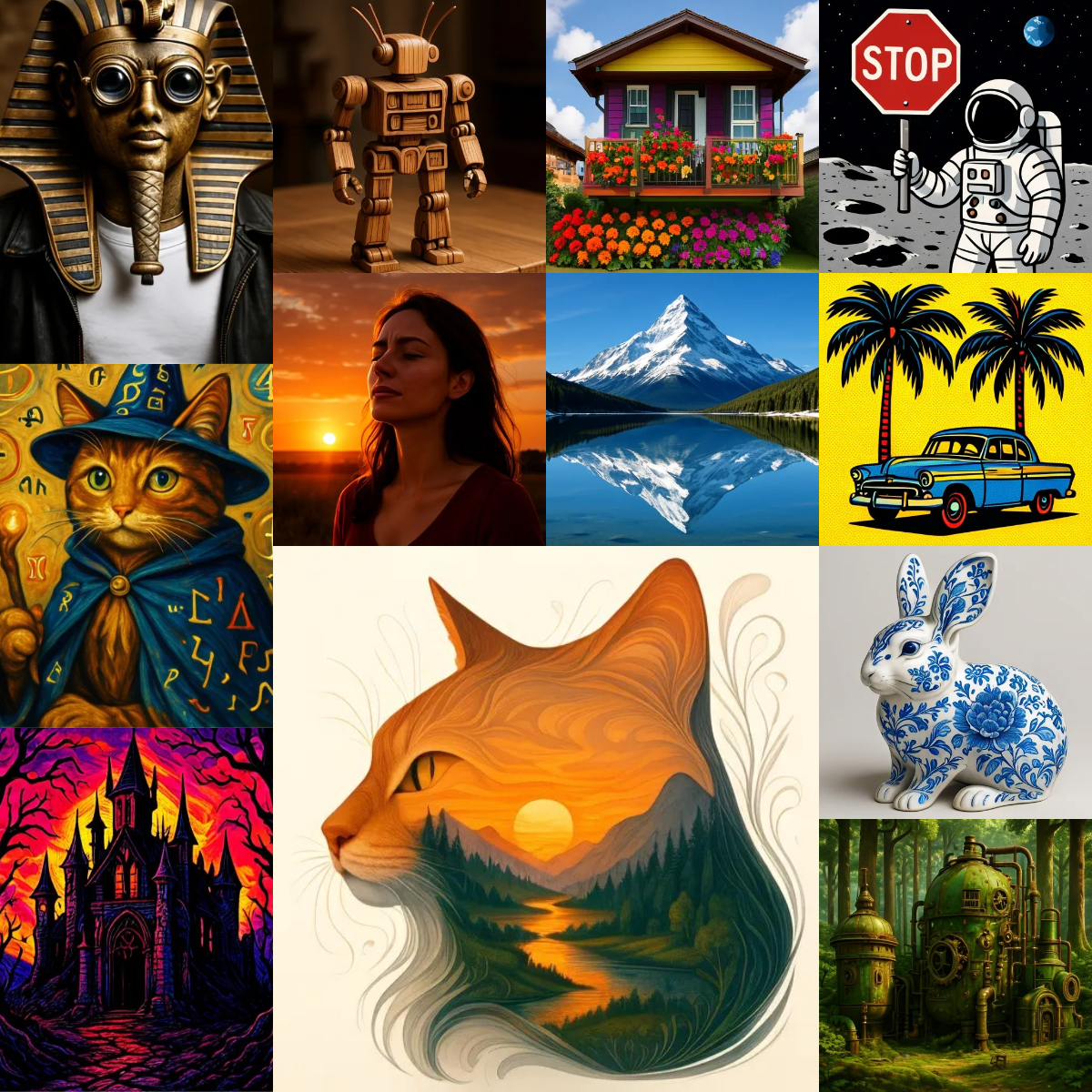}
    \caption{Text-to-image generation results of TBAC-UniImage}
    \label{fig:teaser}
\end{figure}

\newpage

\section{Introduction}
The unification of generative and understanding capabilities in a single model has emerged as one of the most compelling frontiers in the contemporary landscape of AI research. Given the prominent understanding capabilities of current LLMs, the primary inquiry in unifying generation and understanding now concerns how to endow them with generative capacity. The main approaches can be categorized as follows: discretizing images to enable the prediction of the next image token via an LLM~\citep{ge2023making, sun2024autoregressive, wang2025simplear, xin2025lumina}, adding a diffusion head to an off-the-shelf LLM for pixel-level generation~\citep{dong2023dreamllm, tong2024metamorph, pan2025transfer, chen2025blip3, xie2025show}, and fully integrating the training of understanding and generation within a single, unified LLM architecture~\citep{deng2025emerging, liao2025mogao}.

Although the aforementioned methodologies have achieved promising results in both understanding and generation, the debate over which approach is most suitable for such a unified model is far from settled. The autoregressive paradigm, which extends next-token prediction to the image domain, treats an image as a one-dimensional sequence. However, this approach of sequentially predicting the next image token violates the inherent 2-D structure of an image, making it challenging for the model to learn effectively. As for the second type of method—adding a diffusion head to an LLM—it simply combines understanding and generation in a sequential manner, treating the generative component as a tool rather than part of a truly unified model. Finally, developing a single unified model from scratch requires an immense amount of resources, rendering it unaffordable for most researchers.

A key challenge lies in creating a single model that inherits both the understanding prowess of pre-trained MLLMs and the generation fidelity of DiTs, all while maintaining a low training budget. Inspired by Mixture-of-Transformers~\citep{liang2024mixture, deng2025emerging}, we propose TBAC-UniImage, a model that treats a pre-trained MLLM~\citep{liu2023visual} and a Diffusion Transformer (DiT)~\citep{peebles2023scalable} as synergistic components, integrating them through a deep, layer-wise alignment to create a training-friendly architecture. Following MetaQuery~\citep{pan2025transfer} and BLIP-3o \citep{chen2025blip3}, our method involves inserting several learnable queries into the MLLM's input. These queries are subsequently processed through the MLLM, and their resulting hidden states are used as conditions for the DiT. A key innovation of our model is that these conditions are extracted from different layers of the MLLM, rather than simply adopting the final hidden states of MLLM. Specifically, for an MLLM with $m$ layers and a DiT with $n$ layers (where $m \geq n$), the hidden states from layer $m-n$ of the MLLM are fed into the first layer of the DiT, the states from layer $m-n+1$ are fed into the second, and so on, up to the final layers of each model. Drawing an analogy to Ladder-Side Tuning in LLMs \citep{sung2022lst}, we term our layer-wise tuning as \textbf{Ladder-Side Diffusion Tuning}, a method where the diffusion model is fine-tuned and functions as a generative ladder attached to the MLLM. This design allows DiT absorbing information within deeper layers of MLLM, achieving better synergy between understanding and generation. Furthermore, the MLLM and DiT layers to be functionally merged during inference, transforming two separate training models into a single, unified architecture for generation.

Our model demonstrates strong performance across several benchmarks. It achieves a score of 0.87 on GenEval~\citep{ghosh2023geneval}, comparable to the state-of-the-art Qwen-Image~\citep{wu2025qwen}. Furthermore, it obtains a score of 80.97 on DPG-Bench~\citep{hu2024ella}, outperforming both MetaQuery and BLIP-3o. On the recent TIIF-Bench~\citep{wei2025tiif}, our model scores 62.37 on the overall-short subset and 61.18 on the overall-long subset. Notably, within the challenging "Advanced Following" subset of TIIF-Bench, our model achieves scores of 62.91 (short) and 65.03 (long), showcasing highly competitive performance against other leading text-to-image models. Also, our model gets 3.25 on ImgEdit~\citep{ye2025imgedit} benchmark, demonstrating its competitive capability in Image Editing task.

\section{Methodology}

\begin{figure}
    \centering
    \includegraphics[width=0.8\linewidth]{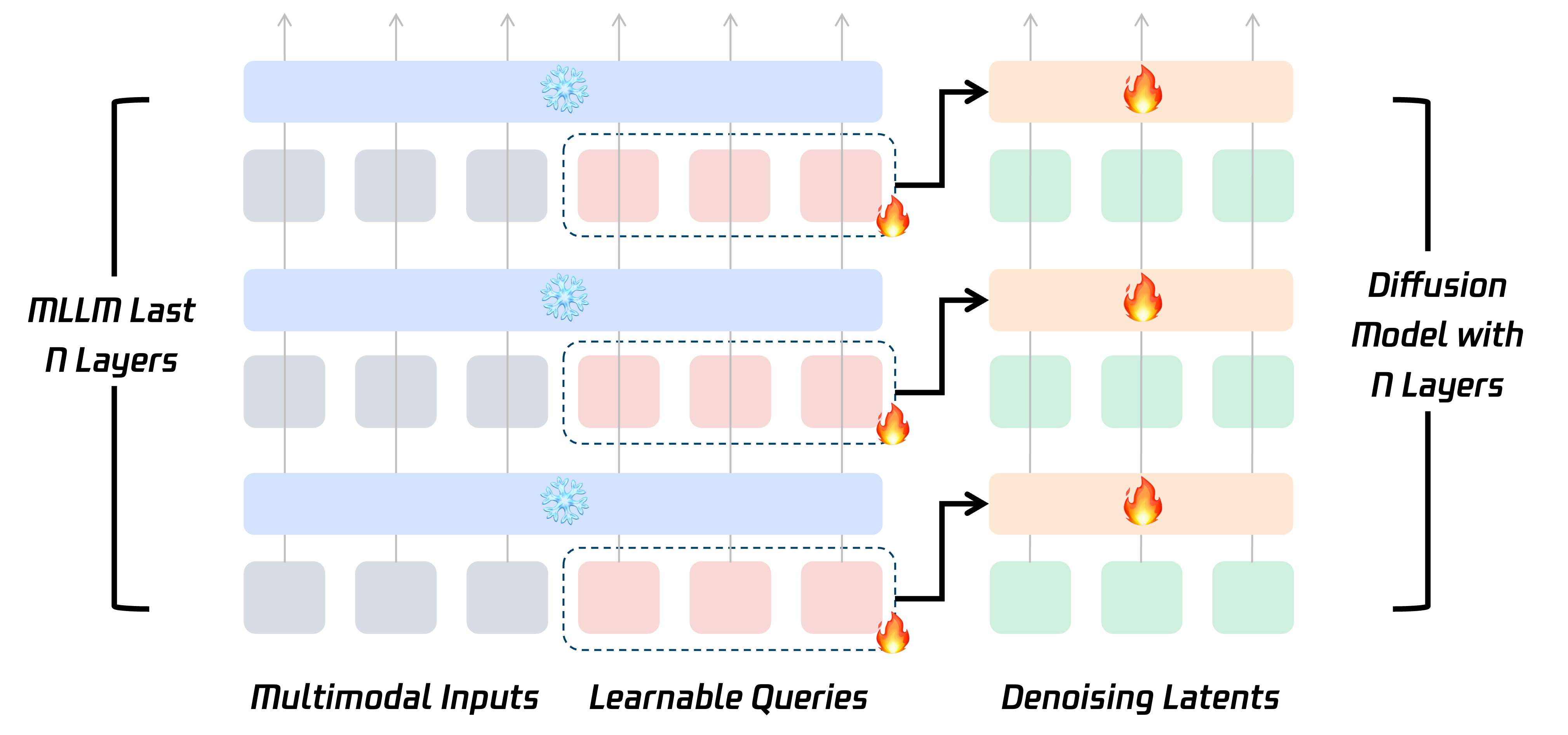}
    \caption{The overview of TBAC-UniImage, where the MLLM parameters are frozen, learnable queries and DiT are tuned together for denoising objective.}
    \label{fig:fig1}
\end{figure}

\subsection{Architecture}
Our proposed model, TBAC-UniImage, establishes a novel interface between a pre-trained, $m$-layer Multimodal Large Language Model (MLLM) and an $n$-layer Diffusion Transformer (DiT). The core of our method is a layer-wise conditioning mechanism facilitated by a set of learnable queries, $\mathcal{Q} \in \mathbb{R}^{N\times D}$. These queries are injected into the MLLM along with the primary input conditions (e.g., text or image-text prompts) and are processed through its layers. We then progressively channel the evolving representations of these queries from the MLLM to the DiT. Formally, for each $(m - n + i)$, our model feed the query hidden states $Q^{(m - n + i)}$ from the $(m - n + i)$-th MLLM layer into a lightweight two-layer connector $f$. This connector, which has only $10M$ trainable paramters, processes the states to generate the conditional input for the $i$-th DiT layer. This creates a deep, hierarchical bridge between the two models. For efficient tuning, we adopt a parameter-efficient training strategy where the MLLM's weights remain frozen, preserving its powerful semantic and multimodal understanding. Training is thus focused exclusively on optimizing the learnable queries $Q$ and the parameters of the DiT. The overview can be found in \cref{fig:fig1}

\subsection{Training Recipe}

In TBAC-UniImage, we use Qwen2.5-VL-3B-Instruct~\citep{bai2025qwen2} as MLLM, SANA-1600M-512px~\citep{xie2024sana} as DiT, the denoising objective is Flow Matching~\citep{lipman2022flow}. The training methodology comprises three sequential stages, which can be found in ~\cref{ref:table1}. The initial stage is text-to-image pre-training, where the diffusion model is trained on text-image pairs. The primary objective of this stage is to align the model's conditional latent space with the high-level feature representations of a MLLM. Subsequently, to leverage the MLLM's intrinsic multimodal understanding capabilities, we introduce a second stage of image-text-to-image pre-training. This phase aims to endow the diffusion model with the capacity for image generation conditioned on a unified multimodal representation. Finally, the model undergoes a fine-tuning stage using a curated dataset, referred to as GPT Image 1~\citep{gptimage} distilled data, which contains a mixture of both text-to-image generation and image editing examples. This concluding step is crucial for significantly enhancing the instruction-following ability and overall quality of the generated outputs.

During the initial phase of the first stage, namely text-to-image pre-training, the training process is prone to instability. Specifically, the gradient norm can experience sudden spikes. To mitigate this phenomenon, we adopt the loss and gradient norm monitoring mechanism from MetaQueries. This involves continuous monitoring of the gradient norm and loss; should its value exceed a predefined threshold (indicating a spike), the optimizer is programmed to discard the current gradients and skip the weight update for that step by zeroing out gradient. This approach effectively prevents catastrophic training failure and promotes stability during the critical early stages of learning.
\begin{table}[hbp]
    \centering
    \caption{Detailed training recipe in three training stages.}
    \label{ref:table1}
    \begin{tabular}{lccc}
        \toprule
        \textbf{Setting} & \textbf{T2I Pretrain} & \textbf{TI2I Pretrain} & \textbf{FT}\\
        \midrule
        Dataset & \begin{tabular}[c]{@{}c@{}}BLIP3o-Long-Caption \\ BLIP3o-JourneyDB \\ \citep{chen2025blip3} \end{tabular} & \makecell{GPT-Image-Edit-1.5M \\ \citep{wang2025gpt}} & \begin{tabular}[c]{@{}c@{}}BLIP3o-60k \\ \citep{chen2025blip3}\\ ShareGPT-4o-Image \\ \citep{chen2025sharegpt} \end{tabular} \\
        \midrule
        Data Size & 30M & 1.5M & 150K \\
        Batch Size & 512 & 256 & 256  \\
        Training Steps & 150K & 60K & 60K  \\
        \midrule
        \multicolumn{4}{l}{\textit{Common Settings for All Stages}} \\
        \midrule
        % Learning Rate & \multicolumn{3}{c}{$10^{-4}$} \\
        % Optimizer & \multicolumn{3}{c}{AdamW} \\
        % Warm-up Steps & \multicolumn{3}{c}{5,000} \\
        Learning Scheduler & Cosine & Optimizer & AdamW \\
        Max Learning Rate & $10^{-4}$ & Min Learning Rate & $10^{-5}$ \\
        Warm-up Steps & 5,000 & Num of Learnable Queries & 64 \\
        \bottomrule
    \end{tabular}
\end{table}

\begin{table}[t]\centering
\caption{Quantitative evaluation results on GenEval.}
\begin{adjustbox}{width=\textwidth}
\begin{tabular}{l|cccccc|c}
\toprule
\multirow{2}{*}{\textbf{Model}} & \textbf{Single} & \textbf{Two} & \multirow{2}{*}{\textbf{Counting}} & \multirow{2}{*}{\textbf{Colors}} & \multirow{2}{*}{\textbf{Position}} & \textbf{Attribute} & \multirow{2}{*}{\textbf{Overall$\uparrow$}} \\
& \bf Object & \bf Object & & & & \bf Binding & \\
\midrule
PixArt-$\alpha$~\citep{chen2024pixartalpha}         & 0.98          & 0.50        & 0.44     & 0.80    & 0.08     & 0.07              & 0.48     \\
SD3 Medium~\citep{esser2024scaling}      & 0.98          & 0.74       & 0.63     & 0.67   & 0.34     & 0.36              & 0.62     \\
FLUX.1 [Dev]~\citep{flux2024}   & 0.98 & 0.81 & 0.74  & 0.79 & 0.22   & 0.45 & 0.66     \\
SD3.5 Large~\citep{esser2024scaling}     & 0.98          & 0.89       & 0.73     & 0.83   & 0.34     & 0.47              & 0.71     \\
Lumina-Image 2.0~\citep{qin2025lumina} & - & 0.87 & 0.67 & -      & - & 0.62 & 0.73     \\
HiDream-I1-Full~\citep{cai2025hidream}          & 1.00             & 0.98       & 0.79     & 0.91   & 0.60      & 0.72              & 0.83     \\
Seedream 3.0~\citep{gao2025seedream} & 0.99 & 0.96 & 0.91 & 0.93 & 0.47 & 0.80 &0.84 \\
GPT Image 1 [High]~\citep{gptimage}           & 0.99          & 0.92       & 0.85     & 0.92   & 0.75     & 0.61              & 0.84     \\
Qwen-Image~\citep{wu2025qwen} & 0.99 & 0.92 & 0.89 & 0.88 &  0.76 & 0.77 & 0.87 \\
Qwen-Image-RL~\citep{wu2025qwen}   & 1.00 & 0.95  & 0.93  & 0.92 & 0.87 & 0.83 & 0.91 \\
% \midrule
Emu3-Gen~\citep{wang2024emu3} & 0.98 & 0.71 & 0.34 & 0.81 & 0.17 &0.21 & 0.54 \\
Show-o~\citep{xie2024show} & 0.95 & 0.52 & 0.49 & 0.82 & 0.11 & 0.28 &0.53 \\
JanusFlow~\citep{ma2025janusflow} &0.97 & 0.59 & 0.45 & 0.83 & 0.53 & 0.42 & 0.63 \\
Janus-Pro-7B~\citep{chen2025janus}     & 0.99          & 0.89       & 0.59     & 0.90    & 0.79     & 0.66              & 0.80      \\
MetaQuery-L~\citep{pan2025transfer}     & -         & -       & -    & -    & -     & -              & 0.78 \\
BLIP3-o-8B~\citep{chen2025blip3}     & -         & -       & -    & -    & -     & -              & 0.83 \\
OpenUni-B-512~\citep{wu2025openuni} & 0.99 & 0.91 & 0.74 & 0.90 & 0.77 & 0.73 \\
Tar-7B~\citep{han2025vision} & - & 0.92 & 0.83 & 0.65 & - & - & 0.83 \\
\midrule
TBAC-UniImage-3B & 0.99 & 0.94 & 0.77 & 0.92 & 0.83 & 0.75 & 0.87 \\
\bottomrule
\end{tabular}\label{tab:geneval}
\end{adjustbox}
\end{table}

\begin{table}[t]
\centering
\caption{Quantitative evaluation results on DPG.}
\small
\begin{adjustbox}{width=\textwidth}
\begin{tabular}{l|ccccc|c}
\toprule
\textbf{Model}           & \bf Global & \bf Entity & \bf Attribute & \bf Relation & \bf Other & \bf Overall$\uparrow$ \\
\midrule
PixArt-$\alpha$~\citep{chen2024pixartalpha}         & 74.97  & 79.32  & 78.60      & 82.57    & 76.96 & 71.11    \\
Lumina-Next~\citep{zhuo2024luminanext}      & 82.82  & 88.65  & 86.44     & 80.53    & 81.82 & 74.63    \\
Playground v2.5~\citep{li2024playground}  & 83.06  & 82.59  & 81.20      & 84.08    & 83.50  & 75.47    \\
Hunyuan-DiT~\citep{li2024hunyuandit}      & 84.59  & 80.59  & 88.01     & 74.36    & 86.41 & 78.87    \\
PixArt-$\Sigma$~\citep{chen2024pixartsigma}         & 86.89  & 82.89  & 88.94     & 86.59    & 87.68 & 80.54    \\
DALL-E 3~\citep{openai2023dalle3}         & 90.97  & 89.61  & 88.39     & 90.58    & 89.83 & 83.50     \\
FLUX.1 [Dev]~\citep{flux2024}       & 74.35  & 90.00     & 88.96     & 90.87    & 88.33 & 83.84    \\
SD3 Medium~\citep{esser2024scaling}       & 87.90   & 91.01  & 88.83     & 80.70     & 88.68 & 84.08    \\
HiDream-I1-Full~\citep{cai2025hidream}          & 76.44  & 90.22  & 89.48     & 93.74    & 91.83 & 85.89    \\
Lumina-Image 2.0~\citep{qin2025lumina} & -      & 91.97  & 90.20      & 94.85    & -     & 87.20     \\
Seedream 3.0~\citep{gao2025seedream} &  94.31     & 92.65  & 91.36      & 92.78    & 88.24     & 88.27     \\
GPT Image 1 [High]~\citep{gptimage} &  88.89     & 88.94  & 89.84      & 92.63    & 90.96     & 85.15     \\
Qwen-Image~\citep{wu2025qwen}& 91.32 & 91.56 & 92.02 & 94.31 & 92.73 & 88.32 \\
% \midrule
Emu3-Gen~\citep{wang2024emu3}         & 85.21  & 86.68  & 86.84     & 90.22    & 83.15 & 80.60     \\
Janus-Pro-1B~\citep{chen2025janus}     & 87.58  & 88.63  & 88.17     & 88.98    & 88.30  & 82.63    \\
Janus-Pro-7B~\citep{chen2025janus}     & 86.90   & 88.90   & 89.40      & 89.32    & 89.48 & 84.19    \\
MetaQuery-L~\citep{pan2025transfer}     & -   & -   & -      & -    & - & 81.10    \\
BLIP3-o-8B~\citep{chen2025blip3}     & -   & -   & -      & -    & - & 80.73    \\
OpenUni-B-512~\citep{wu2025openuni} & 85.87 & 87.33 & 86.54 & 86.91 & 89.43 & 80.29 \\
Tar-7B~\citep{han2025vision} & - & 88.62 & 88.05 & 93.98 & - & 84.19 \\
\midrule
TBAC-UniImage-3B & 83.52 & 87.94 & 87.80 & 87.17 & 87.02 & 80.97 \\
\bottomrule
\end{tabular}\label{tab:dpg}
\end{adjustbox}
\end{table}

\section{Experiments}
% \subsection{Quantitative Results}
To comprehensively evaluate the performance of our proposed TBAC-UniImage model, we conducted extensive experiments on three prominent benchmarks: GenEval~\citep{ghosh2023geneval}, DPG-Bench~\citep{hu2024ella}, TIIF-Bench~\citep{wei2025tiif}, and ImgEdit~\citep{ye2025imgedit}. The experimental results, summarized in \cref{tab:geneval,tab:dpg,tab:tiif,tab:imgedit}, validate the effectiveness of our approach.

On the GenEval benchmark, designed to assess a model's capacity for generating images with precise compositional attributes (e.g., spatial relations, color, and object counts), TBAC-UniImage achieves a score of 0.87. This result not only surpasses state-of-the-art unified models, such as Tar, but also achieves performance comparable to leading specialized text-to-image models like Qwen-Image. The model's superior performance on this benchmark underscores its proficiency in accurately interpreting and rendering fundamental object-property relationships.

For a more stringent evaluation of fine-grained prompt comprehension, we utilized DPG-Bench, a benchmark comprising 1,000 intricate prompts. Our model obtained a score of 80.97. While this is surpassed by several state-of-the-art specialized text-to-image models (e.g., Qwen-Image~\citep{wu2025qwen}, Seedream 3.0~\citep{gao2025seedream}), it remains highly competitive against leading unified models like BLIP3-o~\citep{chen2025blip3} and MetaQuery~\citep{pan2025transfer}, highlighting its strong standing within its model class.

Furthermore, we evaluated our model on the recent, large-scale TIIF-Bench, which assesses performance across three key aspects: Basic Following, Advanced Following, and Designer. Notably, TBAC-UniImage exhibits exceptional performance in the "Advanced Following" category, achieving scores of 62.91 and 65.03 on short and long prompts, respectively. This outperformance, even against some leading text-to-image models, highlights our model's advanced capability to parse and synthesize complex, dense prompts. This result establishes its robust instruction-following fidelity, particularly among open-source models of a comparable scale.

We further benchmark our model's image editing capabilities using the ImgEdit dataset~\citep{ye2025imgedit}. Diverging from conventional approaches, our model \textbf{does not rely on image VAE representations for generation}. It instead fully exploits multimodal semantics captured in a set of learnable queries. Despite this fundamental design choice, our model achieves a superior overall score of 3.25, exceeding the performance of leading models like BAGEL~\citep{deng2025bagel} and Step1X-Edit~\citep{liu2025step1x}. This accomplishment validates the feasibility of high-fidelity image editing using only the intrinsic features of an MLLM, without external generative priors.

\begin{table}[t]\centering
\caption{Quantitative evaluation results on TIIF Bench testmini.}
\renewcommand{\arraystretch}{1.7} 
\setlength{\tabcolsep}{3pt}

\centering
\begin{adjustbox}{width=\textwidth}
\begin{tabular}{l|cc|cccccccc|cccccccccccc|cc}
\toprule
\multirow{3}{*}{\textbf{Model}}
  & \multicolumn{2}{c|}{\multirow{2}{*}{\textbf{Overall}}}
  & \multicolumn{8}{c|}{\textbf{Basic Following}}
  & \multicolumn{12}{c|}{\textbf{Advanced Following}}
  & \multicolumn{2}{c}{\textbf{Designer}} \\

\cmidrule(lr){4-11} \cmidrule(lr){12-23} \cmidrule(lr){24-25}

% ── 2nd header row ───────────────────────────────────────────────────────────
& & &
  \multicolumn{2}{c}{Avg}                    % Basic Avg
  & \multicolumn{2}{c}{Attribute}
  & \multicolumn{2}{c}{Relation}
  & \multicolumn{2}{c|}{Reasoning}
  & \multicolumn{2}{c}{Avg}                  % Advanced Avg
  & \multicolumn{2}{c}{\makecell{Attribute\\+Relation}}
  & \multicolumn{2}{c}{\makecell{Attribute\\+Reasoning}}
  & \multicolumn{2}{c}{\makecell{Relation\\+Reasoning}}
  & \multicolumn{2}{c}{Style}
  & \multicolumn{2}{c|}{Text}
  & \multicolumn{2}{c}{\makecell{Real\\World}} \\

% ── 3rd header row (short / long for each sub‑category) ──────────────────────
& short & long &          % Overall
  short & long &          % Basic‑Avg
  short & long &          % Attribute
  short & long &          % Relation
  short & long &          % Reasoning
  short & long &          % Advanced‑Avg
  short & long &          % Attribute+Relation
  short & long &          % Attribute+Reasoning
  short & long &          % Relation+Reasoning
  short & long &          % Style
  short & long &          % Text
  short & long            % Real‑World
\\
\midrule

% FLUX.1 [dev]~\citep{flux2024}  &{{71.09}} &71.78 &83.12 &78.65& 87.05 & 83.17 & 87.25 &80.39 &75.01 &72.39 &65.79 &68.54& 67.07 &73.69 &73.84 &73.34 &69.09 &71.59 & 66.67 & 66.67 &43.83 &52.83 &70.72 &71.47 \\
FLUX.1 [Pro]~\citep{flux2024} &67.32 &69.89 &79.08 &78.91 &78.83 &81.33 &82.82 &83.82 &75.57 &71.57 &61.10 &65.37 &62.32 &65.57 &69.84 &71.47 &65.96 &67.72 &63.00 &63.00 &35.83 &55.83 &71.80 &68.80 \\
% DALL-E 3~\citep{openai2023dalle3} &74.96 &70.81 &78.72 &78.50 &79.50 &79.83 &80.82 &78.82 &75.82 &76.82 &73.39 &67.27 &73.45 &67.20 &72.01 &71.34 &63.59 &60.72 &89.66 &86.67 &66.83 &54.83 &72.93 &60.99 \\
SD 3~\citep{esser2024scaling} &67.46 &66.09 &78.32 &77.75 &83.33 &79.83 &82.07 &78.82 &71.07 &74.07 &61.46 &59.56 &61.07 &64.07 &68.84 &70.34 &50.96 &57.84 &66.67 &76.67 &59.83 &20.83 &63.23 &67.34 \\
PixArt-$\Sigma$~\citep{chen2024pixartalpha} &62.00 &58.12 &70.66 &75.25 &69.33 &78.83 &75.07 &77.32 &67.57 &69.57 &57.65 &49.50 &65.20 &56.57 &66.96 &61.72 &66.59 &54.59 &83.33 &70.00 &1.83 &1.83 &62.11 &52.41 \\
Lumina-Next~\citep{zhuo2024luminanext} &50.93 &52.46 &64.58 &66.08 &56.83 &59.33 &67.57 &71.82 &69.32 &67.07 &44.75 &45.63 &51.44 &43.20 &51.09 &59.72 &44.72 &54.46 &70.00 &66.67 &0.00 &0.83 &47.56 &49.05 \\
Hunyuan-DiT~\citep{li2024hunyuandit} &51.38 &53.28 &69.33 &69.00 &65.83 &69.83 &78.07 &73.82 &64.07 &63.32 &42.62 &45.45 &50.20 &41.57 &59.22 &61.84 &47.84 &51.09 &56.67 &73.33 &0.00 &0.83 &40.10 &44.20 \\
Show-o~\citep{xie2024show} &59.72 &58.86 &73.08 &75.83 &74.83 &79.83 &78.82 &78.32 &65.57 &69.32 &53.67 &50.38 &60.95 &56.82 &68.59 &68.96 &66.46 &56.22 &63.33 &66.67 &3.83 &2.83 &55.02 &50.92 \\
LightGen~\citep{wu2025lightgen} &53.22 &43.41 &66.58 &47.91 &55.83 &47.33 &74.82 &45.82 &69.07 &50.57 &46.74 &41.53 &62.44 &40.82 &61.71 &50.47 &50.34 &45.34 &53.33 &53.33 &0.00 &6.83 &50.92 &50.55 \\
SANA 1.5~\citep{xie2025sana1d5}   &67.15 &65.73 &79.66 &77.08 &79.83 &77.83 &85.57 &83.57 &73.57 &69.82 &61.50 &60.67 &65.32 &56.57 &69.96 &73.09 &62.96 &65.84 &80.00 &80.00 &17.83 &15.83 &71.07 &68.83 \\
Infinity~\citep{han2025infinity} &62.07 &62.32 &73.08 &75.41 &74.33 &76.83 &72.82 &77.57 &72.07 &71.82 &56.64 &54.98 &60.44 &55.57 &74.22 &64.71 &60.22 &59.71 &80.00 &73.33 &10.83 &23.83 &54.28 &56.89 \\
Janus-Pro-7B~\citep{chen2025janus} &66.50 &65.02 &79.33 &78.25 &79.33 &82.33 &78.32 &73.32 &80.32 &79.07 &59.71 &58.82 &66.07 &56.20 &70.46 &70.84 &67.22 &59.97 &60.00 &70.00 &28.83 &33.83 &65.84 &60.25 \\
MidJourney v7~\citep{midjourneyv7} &68.74 &65.69 &77.41 &76.00 &77.58 &81.83 &82.07 &76.82 &72.57 &69.32 &64.66 &60.53 &67.20 &62.70 &81.22 &71.59 &60.72 &64.59 &83.33 &80.00 &24.83 &20.83 &68.83 &63.61 \\
Qwen-Image~\citep{wu2025qwen} &86.14 &86.83 & 86.18 & 87.22 & 90.50 & 91.50 & 88.22 & 90.78 & 79.81 & 79.38 & 79.30 & 80.88 & 79.21 &78.94 & 78.85 &81.69 & 75.57 &78.59 & 100.00  &100.00 & 92.76 & 89.14 & 90.30 &91.42 \\
\midrule
\bf TBAC-UniImage-3B & 62.37 & 61.18 & 75.92 & 73.97 & 76.50 & 71.75 & 81.44 & 79.64 & 69.84 & 70.52 & \textbf{62.91} & \textbf{65.03} & 74.65 & 71.61 & 57.47 & 63.65 & 61.99 & 67.59 & 80.00 & 66.67  & 2.71 & 1.36 & 56.72 & 57.84 \\
\bottomrule
\end{tabular}\label{tab:tiif}
\end{adjustbox}
\end{table}

\begin{table}[!t]
    \centering
    \scriptsize
    % \resizebox{0.99\linewidth}{20mm}{
    \caption{Quantitative evaluation results on ImgEdit.}
    \resizebox{0.99\linewidth}{!}{
    \begin{tabular}{l|ccccccccc|c}
        \toprule
        \textbf{Model} & \bf Add & \bf Adjust & \bf Extract & \bf Replace & \bf Remove & \bf Background & \bf Style & \bf Hybrid & \bf Action & \bf Overall $\uparrow$ \\
        \midrule
        MagicBrush~\citep{zhang2023magicbrush} & 2.84 & 1.58 & 1.51 & 1.97 & 1.58 & 1.75 & 2.38 & 1.62 & 1.22 & 1.90 \\
        Instruct-Pix2Pix \citep{brooks2023instructpix2pix} & 2.45 & 1.83 & 1.44 & 2.01 & 1.50 & 1.44 & 3.55 & 1.20 & 1.46 & 1.88 \\
        AnyEdit~\citep{yu2025anyedit} & 3.18 & 2.95 & 1.88 & 2.47 & 2.23 & 2.24 & 2.85 & 1.56 & 2.65 & 2.45 \\
        UltraEdit~\citep{zhao2024ultraedit} & 3.44 & 2.81 & 2.13 & 2.96 & 1.45 & 2.83 & 3.76 & 1.91 & 2.98 & 2.70 \\
        OmniGen~\citep{xiao2025omnigen} & 3.47 & 3.04 & 1.71 & 2.94 & 2.43 & 3.21 & 4.19 & 2.24 & 3.38 & 2.96 \\
        ICEdit~\citep{zhang2025context} & 3.58 & 3.39 & 1.73 & 3.15 & 2.93 & 3.08 & 3.84 & 2.04 & 3.68 & 3.05 \\
        Step1X-Edit~\citep{liu2025step1x} & 3.88 & 3.14 & 1.76 & 3.40 & 2.41 & 3.16 & 4.63 & 2.64 & 2.52 & 3.06 \\
        BAGEL~\citep{deng2025bagel} & 3.56 & 3.31 & 1.70 & 3.30 & 2.62 & 3.24 & 4.49 & 2.38 & 4.17 & 3.20 \\
        UniWorld-V1~\citep{lin2025uniworldv1} & 3.82 & 3.64 & 2.27 & 3.47 & 3.24 & 2.99 & 4.21 & 2.96 & 2.74 & 3.26 \\
        OmniGen2~\citep{wu2025omnigen2} & 3.57 & 3.06 & 1.77 & 3.74 & 3.20 & 3.57 & 4.81 & 2.52 & 4.68 & 3.44 \\
        % FLUX.1 Kontext [Pro]~\citep{labs2025kontext} & 4.25 & 4.15 & 2.35 & 4.56 & 3.57 & 4.26 & 4.57 & 3.68 & 4.63 & 4.00 \\
        GPT Image 1 [High]~\citep{gptimage} & 4.61 & 4.33 & 2.90 & 4.35 & 3.66 & 4.57 & 4.93 & 3.96 & 4.89 & 4.20 \\
        Qwen-Image~\citep{wu2025qwen} & 4.38 & 4.16 & 3.43 & 4.66 & 4.14 & 4.38 & 4.81 & 3.82 & 4.69 & 4.27 \\
        \midrule
        TBAC-UniImage-3B & 3.79 & 3.20 & 1.59 & 3.85 & 2.54 & 3.91 & 4.52 & 2.38 & 3.52 & 3.25 \\
        \bottomrule
    \end{tabular}
    }
    \label{tab:imgedit}
\end{table}

% \subsection{Qualitative Results}

\begin{figure}
    \centering
    \includegraphics[width=0.9\linewidth]{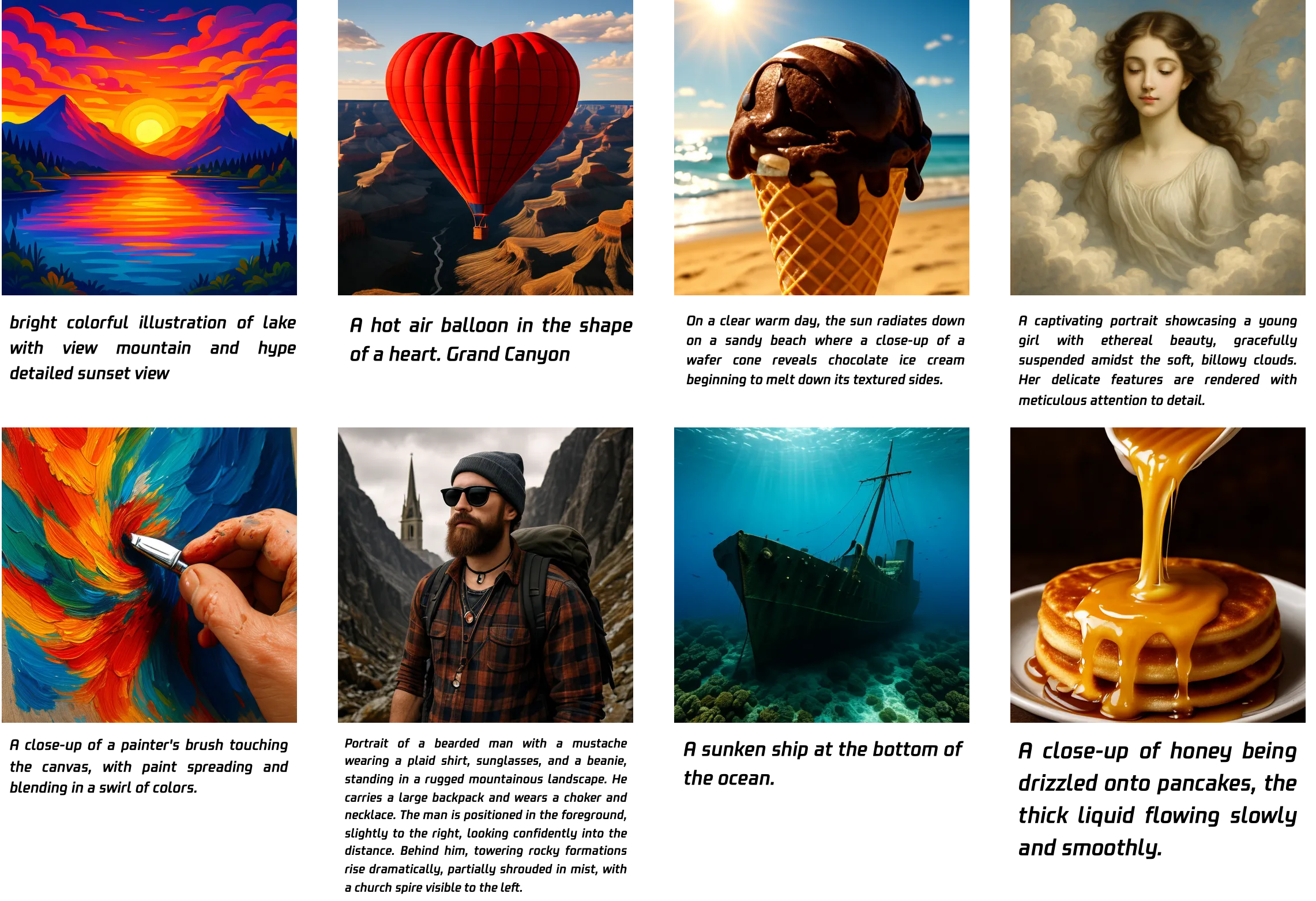}
    \caption{Text-to-image generation samples generated by TBAC-UniImage.}
    \label{fig:fig2}
\end{figure}

\begin{figure}
    \centering
    \includegraphics[width=0.9\linewidth]{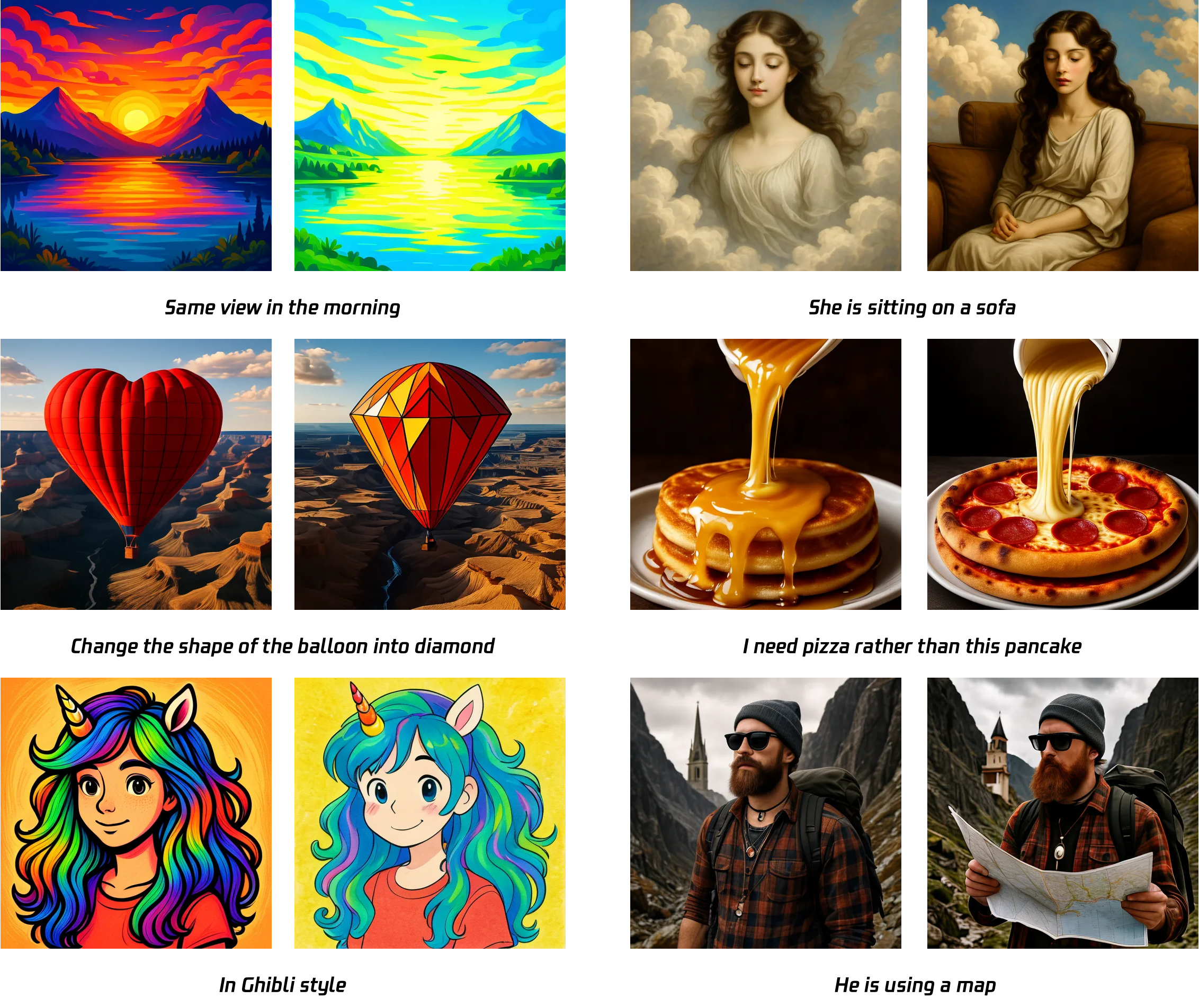}
    \caption{Image Editing samples generated by TBAC-UniImage.}
    \label{fig:fig3}
\end{figure}

\section{Conclusion}
This paper introduces TBAC-UniImage, a novel architecture designed to unify multimodal understanding and generation. Our approach proposes a direct link between a Multimodal Large Language Model (MLLM), which excels at understanding, and a Diffusion Transformer (DiT), a powerful generative model. The core idea of integration is our proposed \textbf{Ladder-Side Diffusion Tuning} mechanism. This technique treats the DiT as a generative ladder for the MLLM, channeling hidden states from various MLLM layers to the DiT. These states serve as rich, multi-level conditional inputs, guiding the image generation process with nuanced understanding. Extensive evaluations on four widely-used benchmarks demonstrate the efficacy of our method. Nevertheless, we acknowledge several limitations that present avenues for future work: (1) enhancing the model's fine-grained comprehension of complex, dense prompts; (2) improving generation consistency for image editing tasks; and (3) refining the in-image text rendering capability. In conclusion, we posit that TBAC-UniImage and our Ladder-Side Diffusion Tuning paradigm offer a valuable new perspective for the research and development of next-generation unified models.

\bibliographystyle{colm2024_conference}
\bibliography{colm2024_conference}

\end{document}

%% file: math_commands.tex
\usepackage{amsmath,amsfonts,bm}

\def\eqref#1{equation~\ref{#1}}
\def\1{\bm{1}}

\DeclareMathAlphabet{\mathsfit}{\encodingdefault}{\sfdefault}{m}{sl}
\SetMathAlphabet{\mathsfit}{bold}{\encodingdefault}{\sfdefault}{bx}{n}